\documentclass[letterpaper, 10 pt, conference]{ieeeconf}  

\IEEEoverridecommandlockouts                              

\usepackage{formatting} 
\usepackage{multirow} 

\usepackage{hyperref}

\title{\LARGE \bf
DenseTact-Mini: An Optical Tactile Sensor for Grasping Multi-Scale Objects From Flat Surfaces
}

\author{Won Kyung Do, Ankush Kundan Dhawan, Mathilda Kitzmann, and Monroe Kennedy III 
\thanks{Authors are members of the ARMLab in the Mechanical Engineering Department, Stanford University, Stanford, CA 94305, USA. 
{\tt\small \{wkdo, ankushd, mathilda, monroek\}@stanford.edu.} 
The first author is supported by a fellowship from the Kwanjeong Educational Foundation. This work is supported by the National Science Foundation under Grants 2142773 and 2220867. Project website with videos are available here: \href{https://sites.google.com/view/densetact-mini/home}{https://sites.google.com/view/densetact-mini/home}   } }

\begin{document}

\maketitle
\thispagestyle{empty}
\pagestyle{empty}

\begin{abstract}

Dexterous manipulation, especially of small daily objects, continues to pose complex challenges in robotics. This paper introduces the DenseTact-Mini, an optical tactile sensor with a soft, rounded, smooth gel surface and compact design equipped with a synthetic fingernail. We propose three distinct grasping strategies: tap grasping using adhesion forces such as electrostatic and van der Waals, fingernail grasping leveraging rolling/sliding contact between the object and fingernail, and fingertip grasping with two soft fingertips. Through comprehensive evaluations, the DenseTact-Mini demonstrates a lifting success rate exceeding 90.2\% when grasping various objects, spanning items from 1 mm basil seeds and small paperclips to items nearly 15mm. This work demonstrates the potential of soft optical tactile sensors for dexterous manipulation and grasping.

\end{abstract}

\section{Introduction}

To enable seamless robot-human collaborations within shared environments, dexterous manipulation is critical. While humans effortlessly grasp and manipulate objects of various shapes and sizes, robots struggle with these tasks, especially when it comes to smaller items. Research has sought to address these challenges, presenting solutions ranging from innovative manipulation strategies and harnessing tactile feedback to pioneering new gripper designs specifically for handling small-objects.

Tactile sensing through vision-based approaches is a promising avenue, offering rich contact information and the potential for enhanced dexterous manipulation. Yet, certain nuances, like the hardness of the contact surface and the synergy between the sensor and the gripper's shape, have been somewhat understudied. Most research has largely emphasized how tactile sensors `detect' objects, particularly small ones. However, how these sensors effectively `grasp' everyday minuscule items has not been thoroughly explored. Conventional optical tactile sensors have rigid gel builds, limiting sensing capabilities and blocking diverse manipulation strategies that are possible with softer materials. Inspired by the sensory fingertip and the rigid nails of human fingers, we propose a new optical tactile sensor with soft, rounded gel surface and fingernail design to facilitate a broader range of grasping strategies.

Our paper's contributions include: 1) A novel, compact tactile sensor that exhibits an ultra-soft, rounded gel surface, making it adept at grasping especially small objects, 2) The strategic integration of a fingernail design on the DenseTact-Mini, enhancing its capability to handle thin, small objects, and 3) An exhaustive exploration and evaluation of various grasping strategies for picking up objects of different dimensions from a flat surface.

The remainder of this paper is composed as follows: Section \ref{sec:related} displays related works, Section \ref{sec:dtmini} describes the fabrication and design process of the DenseTact-Mini, Section \ref{sec:strategy} proposes the grasping methodologies for various objectsizes, Section \ref{sec:evaluation} evaluates each grasping strategy, and Section \ref{sec:conclusion} discusses the conclusions and future work.

\begin{figure}[t]
\vskip 0.04in
\begin{center}
\centerline{\includegraphics[width=\columnwidth]{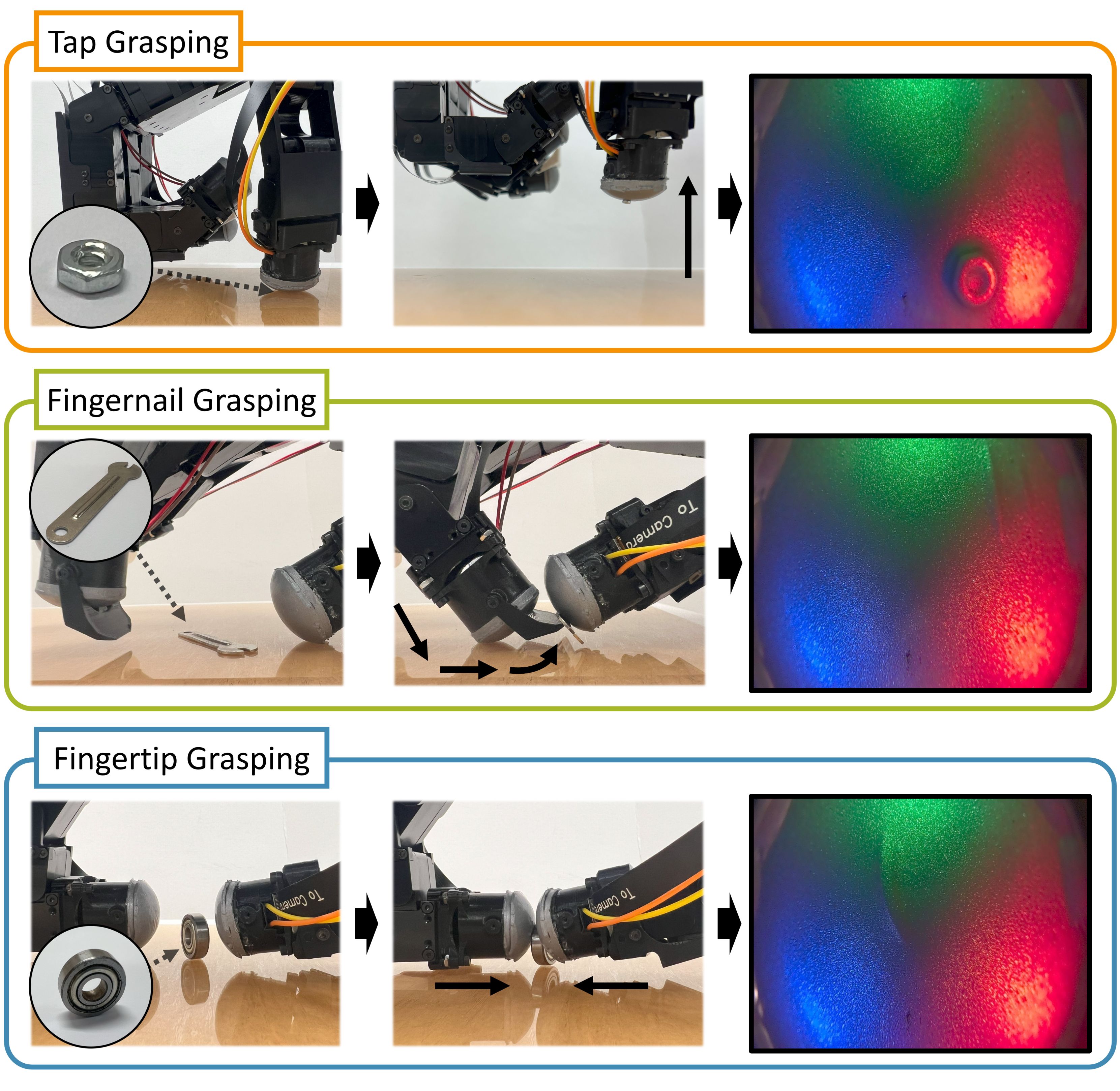}}
\caption{\textbf{DenseTact-Mini.} Grasping various objects with DenseTact-Mini using tap, fingernail, and fingertip strategies.}
\label{fig:main}
\end{center}
\vskip -0.35in
\end{figure}

\section{Related Works} \label{sec:related}

Since dexterity similar to that of human fingers is often required for precise manipulation tasks, robotic grippers of various operating principles have been explored. Soft-robotic hand-like grippers have been proposed to pick up objects of various sizes, largely depending on the size and dimension of the gripper design  \cite{Li2022ReviewPaper, terrile2021gripperreview}. Some gripper designs lean on van der Waals forces for specifically picking up micro-scale objects \cite{Safaric2014VdWpaper}, or maximizing them in a gecko-inspired fashion \cite{cutkosky2018gecko}. Such grippers are capable of gripping either tennis ball-sized or micro-sized objects, but fail to grasp smaller and flatter everyday objects. 

Manipulation of smaller objects has been extensively explored from multiple perspectives. Specialized grippers that precisely handle small, flat, and thin objects include grippers with diverse grasping modes \cite{watanabe2021variable}, digging grippers suited for cluttered environments \cite{zhao2022learning}, and grippers with retractable fingernails \cite{jain2023nailgripper}. Concurently, numerous strategies have been introduced for grasping objects on flat surfaces with two fingers. Many of these strategies are tailored to specialized grippers, which can limit the generalizability of the grasping task \cite{odhner2012precision, hang2019pre}. Furthermore, research focusing on general grippers has often concentrated on grasping larger, flat objects \cite{sarantopoulos2018human, babin2019stable, chavan2020planar}.

Tactile sensors, notably vision-based, have been studied for enhancing dextrous manipulation for their ability to provide rich contact information \cite{do2022densetact, do2023densetact, yuan2017gelsight,Ward2018TacTip,  lambeta2020digitpaper}, useful for in-hand manipulation, classification, or tactile exploration \cite{qi2023general, do2023inter, Dai2022halleffect, Taylor2022gelslim, solano2023embedded}. However, no existing gripper has the combined ability to grasp and sense small and flat daily objects in a generalized fashion. This presents the need for exploring various grasping strategies of everyday objects using a high-resolution optical tactile sensor adapted for enhanced generalized manipulation via an attachable fingernail. 


\begin{figure}[t]
\vskip 0.04in
\begin{center}
\centerline{\includegraphics[width=\columnwidth]{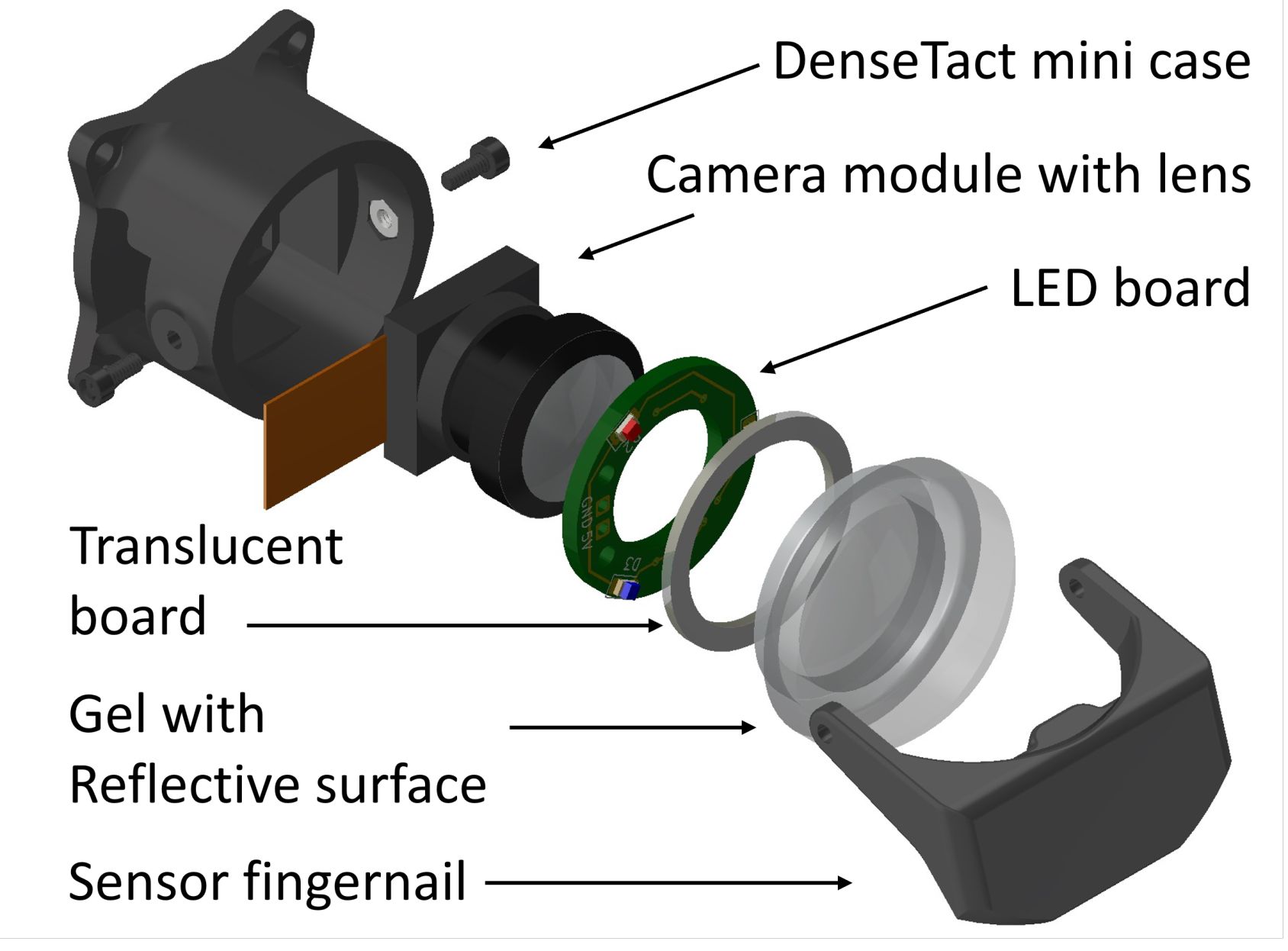}}
\caption{\textbf{Exploded view} of DenseTact-Mini with fingernail.}
\label{fig:explode}
\end{center}
\vskip -0.4in
\end{figure}

\section{DenseTact-Mini} \label{sec:dtmini}

Tactile sensors, especially optical variants, are useful for robot-environment interactions during tasks such as object manipulation. However, grasping small objects remains difficult due to integration challenges and the curse of dimensionality of contact situation. To tackle these, we present a tactile sensor with three specialized grasping strategies. Our sensor has a compact 24mm size with a 3D contoured surface, fitting various high degrees of freedom (DOF) grippers. It is outfitted with a fingernail component for grasping small, flat objects, paired with a soft gel segment for everyday items. Additionally, it features a 60 Hz camera module and fisheye lens to address dynamic contact and grasping situations.

\subsection{Clear gel with camera lens for tactile sensing} \label{sec:gelpart}

To proficiently grasp objects, especially those in the range of 1-2mm, it is imperative that the camera monitors gel deformations upon contact. This means that an object about 1-2mm in size can be observed in around 50 pixels in the camera frame. Furthermore, the gel should be highly curved to enable various grasping strategies from a single sensor.

Figure \ref{fig:explode} presents the exploded view of the DenseTact-Mini. The design is similar to those presented in \cite{do2022densetact, do2023densetact}. The gel's high curvature is generated from a 30mm diameter spherical mold. A fisheye lens with a $222^\circ$ field of view sits 7.6mm below the gel's top surface. The gel hardness is 16 Shore A, and the P-565 silicone base and activator are combined at a $10:1$ ratio. Even though the gel is harder than previous DenseTact sensors, the gel is still softer than other optical tactile sensors such as Gelsight or Digit \cite{yuan2017gelsight, lambeta2020digit}, enabling more effective grasping by maximizing the contact surface between the object and the sensor. STL files detailing both the mold for the gel and its design template are accessible via the project website.
\begin{figure}[t]
\vskip 0.04in
\begin{center}
\centerline{\includegraphics[width=\columnwidth]{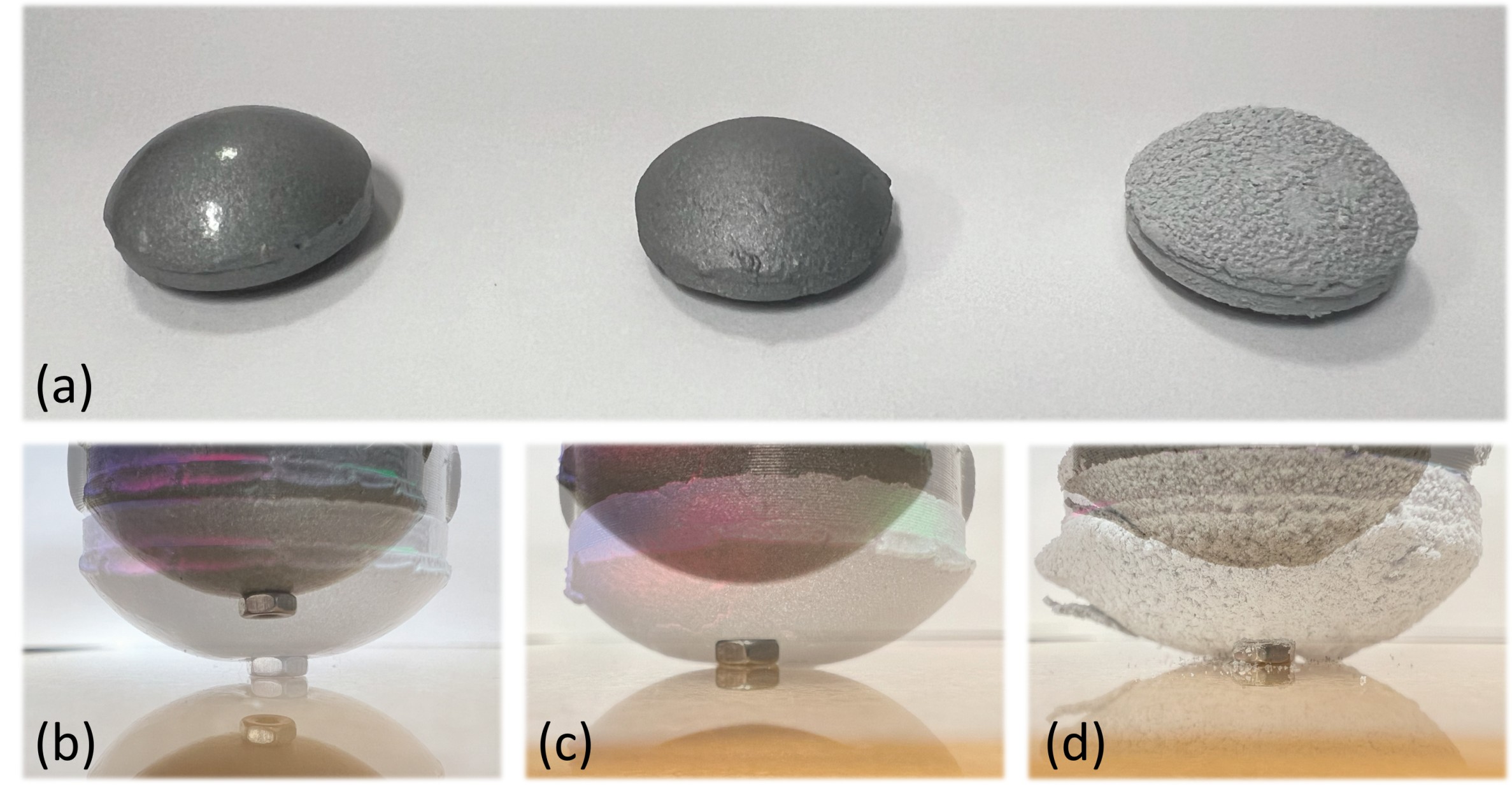}}
\caption{\textbf{Gel for increasing van der Waals force} (a) shows the three different types of gel, and (b)-(d) shows gel's deformation while tap grasping an M2 nut.}
\label{fig:diffsurface}
\end{center}
\vskip -0.4in
\end{figure}

To efficiently grasp small objects and shield the gel from external light, a reflective layer was added after the gel cured. Figure \ref{fig:diffsurface} showcases various gel types. The reflective coating was created by blending metallic ink with Psycho Paint\textsuperscript{TM}, thinned using NOVOCS\textsuperscript{TM} solvent. The final weight ratio for the coating is $\text{Metalic ink} : \text{Base A}: \text{Base B} : \text{NOVOCS solvent} = 19 : 50 : 50 : 100$. After air-spraying the compounded silicone onto the gel, it was cured for 24 hours. Our chosen gel, depicted in Figure \ref{fig:diffsurface} (a), maximizes van der Waals forces between the gel and target objects, explained further in Section \ref{sec:vanderwaals}.

\subsection{Attachable fingernail for enhanced contact manipulation}


Most tactile sensors, when used as on fingertip of a gripper or as the gripper itself (provided they lack a sharp edge), display limited manipulation capabilities due to their sensor design. For instance, a flat sensor struggles to effectively grasp thin, small objects \cite{yuan2017gelsight}. In contrast, round-shaped tactile sensors often require specialized or custom grippers for manipulation \cite{pai2023tactofind, do2023inter, Liu2023gelsighthand}. However, even these grippers find it challenging to grasp small, flat objects from a flat surface, primarily due to 1) the sensor size and 2) the rounded sensor surface shape. To address this limitation, we introduced a fingernail on top of the sensor to enhance grasping performance, especially for flat objects.
\begin{figure}[t]
\vskip 0.04in
\begin{center}
\centerline{\includegraphics[width=\columnwidth]{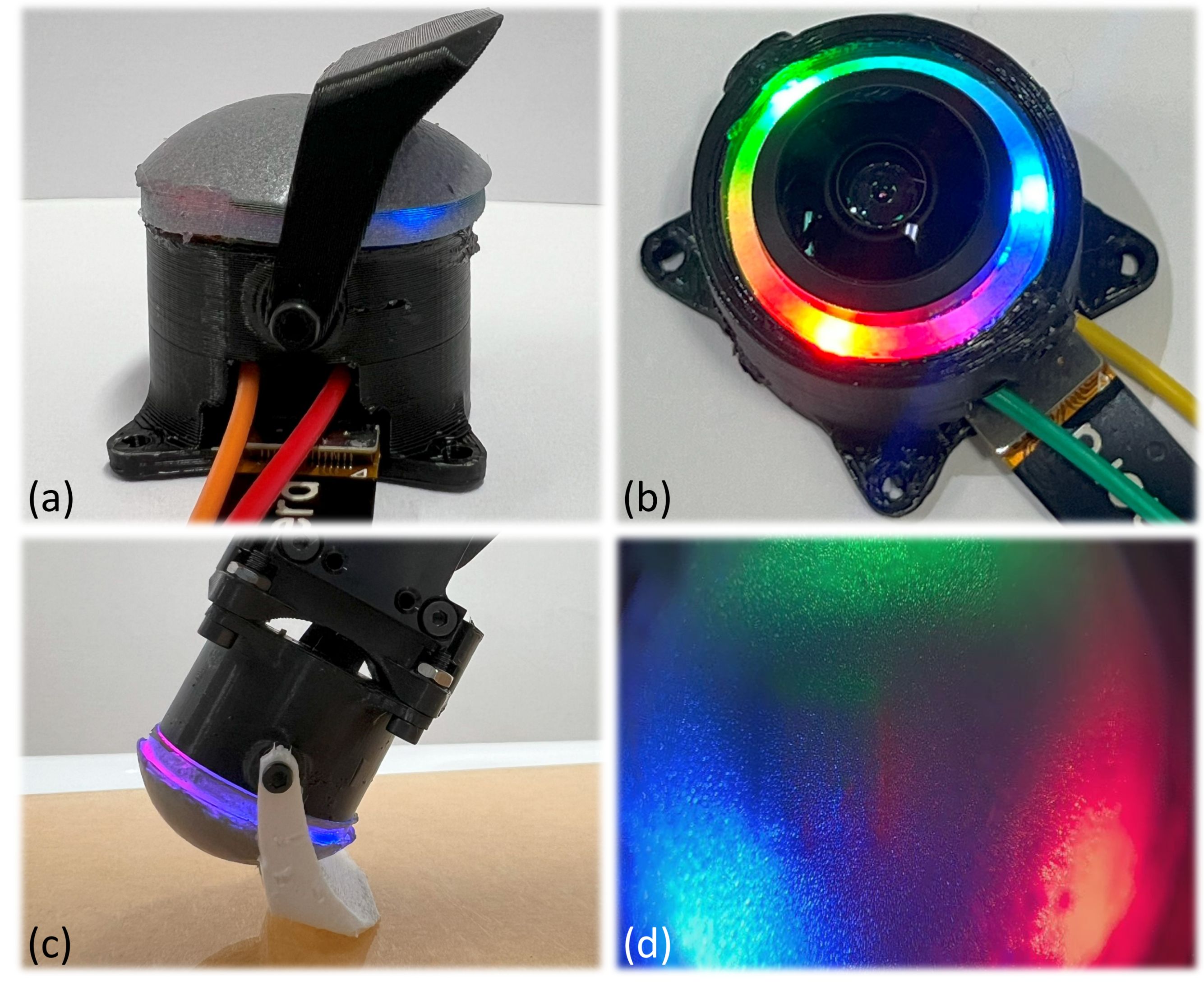}}
\caption{\textbf{DenseTact-Mini Design}: (a) shows the DenseTact-Mini with PLA fingernail, (b) represents the inner view of the sensor, (c) the fingernail is made of TPU, and (d) shows an image taken from DenseTact-Mini with fingernail.}
\label{fig:fingernail_all}
\end{center}
\vskip -0.4in
\end{figure}
The fingernail design is illustrated in Fig. \ref{fig:fingernail_all}. As depicted in (a) of Fig. \ref{fig:fingernail_all} and in Fig. \ref{fig:explode}, the fingernail is positioned atop the sensor and secured to the DenseTact-Mini case using M1.6 screws and nuts. Leveraging the screws as pivot points, the fingernail can passively rotate along one axis. This allows the fingernail to sense torsional force and restrict unnecessary movement during contact. Nevertheless, since the pivot point isn't aligned with the gel's center, the gel obstructs the fingernail's rotational motion after a certain extent. The indented section of the fingernail, adjacent to the gel, creates added friction, further limiting the fingernail's rotation. As a result, the fingernail's movement can be observed through the gel, as demonstrated in (d) of Fig. \ref{fig:fingernail_all}.

The fingernail's pointed edge aids in grasping tiny objects and detects contact. Covering 10.4 degrees of the sensor's 180-degree field of view, the fingernail utilizes 43\% of the gel part for grasping with the fingernail. The rest of the part can also be used for tactile sensing. This 3D-printed component, which can be made from various materials, has a 27.7-degree angle at its tip. While our evaluation employed a polylactic acid (PLA) fingernail, Thermoplastic polyurethane (TPU) provides enhanced flexibility, giving the fingertip a softer touch, as shown in Fig. \ref{fig:fingernail_all}. The STL files are available on the project website.

\subsection{Illumination via LEDs}

A custom PCBA was designed in KiCAD to deliver light to the inner gel surface using parallel red, green, and blue LEDs. Since the luminous intensity of each of the LEDs is different based on color, resistors of varying resistances were added in series with each LED to ensure each LED has the same brightness. When the board is powered with 5V, 79 mA are drawn by the circuit, giving a luminous intensity of 200 mcd for each of the LEDs using the calculated series resistances. A capacitor was added between power and ground to mitigate voltage-step transients. The schematic and pcb files are available on the project website.

To optimize LED light dispersion on the gel's reflective surface, we added a 1mm-thick translucent board above the PCB (see Fig. \ref{fig:explode}). This board, covered with white masking tape, has clear sections 50° from each LED's position. With LEDs at 0°, 120°, and 240°, clearings were at 170°-190°, 290°-310°, and 50°-70°. The final LED setup is in Fig. \ref{fig:fingernail_all}(b).
\begin{table}[t]
\vskip 0.04in
\begin{center}
\begin{tabular}{|c|c|c|c|}
\cline{2-4}
\multicolumn{1}{c}{} & \multicolumn{3}{|c|}{\textbf{\begin{tabular}[c]{@{}c@{}}Gel and Lift Types \\Lift / Short lift / No lift (Success Rate(\%))\end{tabular}}}
\\
\hline
\textbf{Object} & \textbf{\begin{tabular}[c]{@{}c@{}}Gloss Gel \end{tabular}} & \textbf{\begin{tabular}[c]{@{}c@{}}Matte/Gloss Gel\end{tabular}} & \textbf{\begin{tabular}[c]{@{}c@{}}Matte Gel\end{tabular}}  \\ 
\hline
\ Basil Seed      & \begin{tabular}[c]{@{}c@{}}41/9/0 (100)\end{tabular}                                                                 & \begin{tabular}[c]{@{}c@{}}44/6/0 (100)\end{tabular}                                                        & \begin{tabular}[c]{@{}c@{}}48/1/1 (98)\end{tabular}                                                                 \\ \hline 
M1.6 Nut        & \begin{tabular}[c]{@{}c@{}}46/4/0 (100)\end{tabular}                                                              & \begin{tabular}[c]{@{}c@{}}18/29/3 (94)\end{tabular}                                                           & \begin{tabular}[c]{@{}c@{}}1/7/42 (16)\end{tabular}                                                                             \\ \hline 
M2 Nut          & \begin{tabular}[c]{@{}c@{}}46/4/0 (100)\end{tabular}                                                             & \begin{tabular}[c]{@{}c@{}}3/37/10 (80)\end{tabular}                                                            & \begin{tabular}[c]{@{}c@{}}0/1/49 (2)\end{tabular}                                                                              \\ \hline
\end{tabular}
\caption{\textbf{Reflective surface evaluation for adhesion force.} The glossy gel showed the best sustained lift ($>3s$) with minimal short lift ($\leq 3s$) and no lift instances.}
\label{table:gelexperiment}
\end{center}
\vskip -0.3in
\end{table}
\subsection{Sensor assembly}

After fabricating each component, the sensor was assembled by sequentially stacking each module into the sensor's case. Given the modular design of each component, parts can be effortlessly replaced. An IMX219 Camera module's input is processed using an NVIDIA Jetson Orin. This camera works at speeds up to 60Hz, ample for executing contact manipulation tasks in real-time. Moreover, the Jetson module facilitates standalone processing of the sensor's camera feed via a ROS system. The dimensions of the DenseTact-Mini are $24mm \times 26mm \times 24mm$, excluding the fingernail, where the 26mm width corresponds to the direction from which the camera cable extends. With the fingernail attached, the dimensions expand to $25.6mm \times 32.6mm \times 29.3mm$. The sensor's weight, including the fingernail, is 15.35g. The DenseTact-Mini is priced under \$30, with the majority of the cost stemming from the camera (\$25.99), relative to the Gel component (\$0.231), 3D printed parts (\$2), and PCBA (\$1.13).

\section{Grasping strategy of various sizes of an object using DenseTact-Mini}

Three different strategies for grasping small objects are proposed using the DenseTact-Mini. The modelled view of each strategy is depicted in Fig. \ref{fig:main}: tap grasping, fingernail grasping, and fingertip grasping. The tap grasping strategy grasps small, lightweight objects (1mm - 3mm) by tapping the object with the DenseTact-Mini. Fingernail grasping refers to grasping small, flat objects from flat surfaces by sliding the object between two DenseTact-Minis, one with and one without a fingernail. Any flat object with a thin profile can be grasped using this strategy. Finally, the fingertip grasping strategy grasps larger objects with two DenseTact-Minis both without fingernails.

\subsection{Grasping millimeter-scale objects via tap grasping} \label{sec:vanderwaals}

Grasping 1-3mm size of objects using a traditional gripper is challenging due to the gripper's size and the non-negligible adhesion force between the gripper and the object. With DenseTact-Mini's clear gel surface, we can harness the surface interaction forces for a tap grasping strategy. As discussed in \cite{menciassi2004macro, chu2012dual, Tadmor2000LondonVdW}, the adhesion force $F_{adh}$ prevalent in the small object can be represented as:

\begin{align}
F_{adh} &= F_{e} + F_{vdW} + F_{st} \simeq  \frac{\pi \epsilon_0 R_o\sigma^2}{d} + \frac{AR_oR_{dt}}{6(R_o+R_{dt})d^2} \\
\Sigma F &= F_{adh} - mg 
\end{align}

\begin{figure}[t]
\vskip 0.04in
\begin{center}
\centerline{\includegraphics[width=0.97\columnwidth]{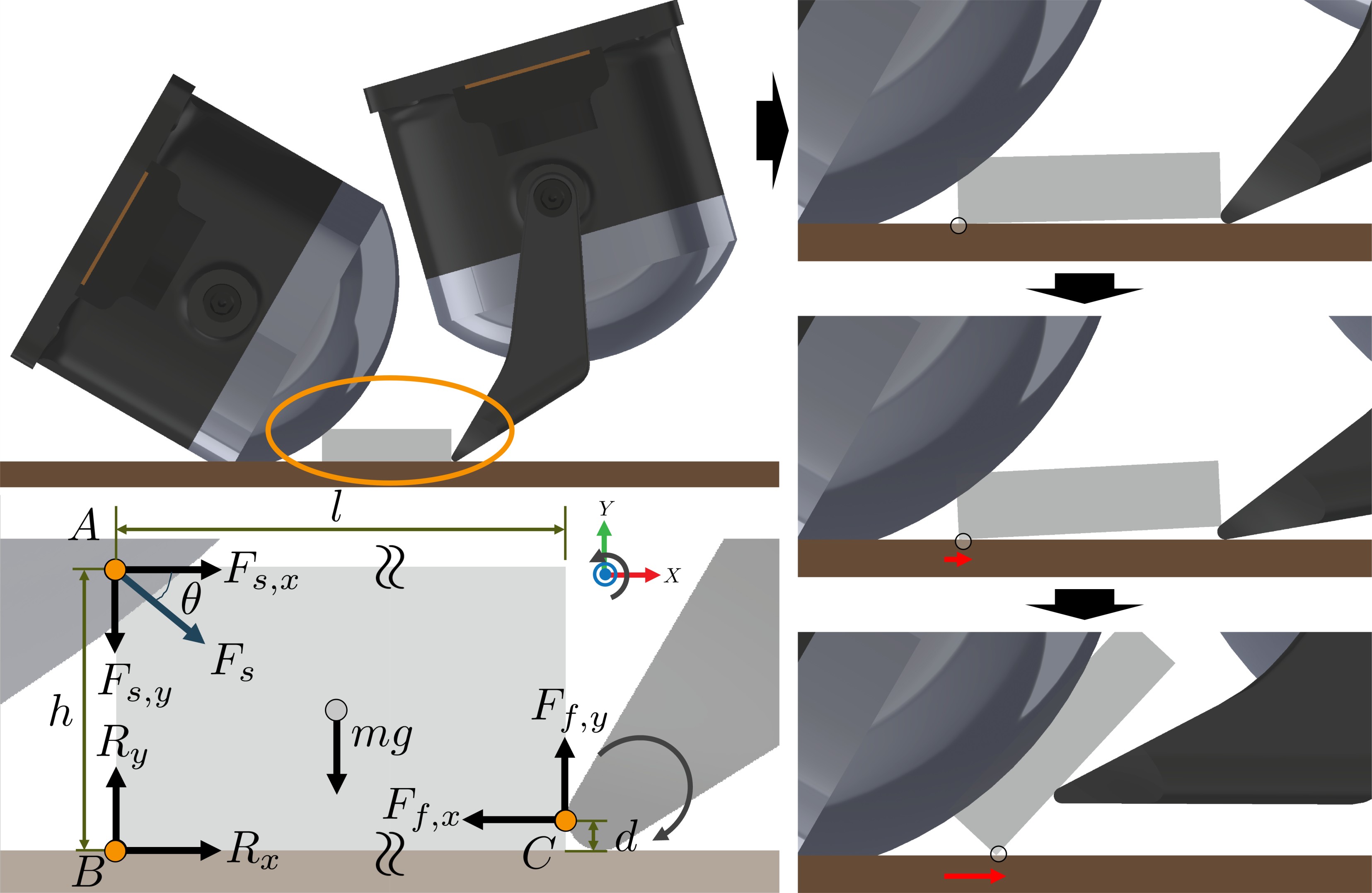}}
\caption{\textbf{Rolling contact grasping strategy.} The fingernail makes rolling contact with the object via a scooping motion.}
\label{fig:fn_strat_1}
\end{center}
\vskip -0.4in
\end{figure}
Where $F_e$, $F_{vdW}$, and $F_{st}$ refer to the electrostatic force, van der Waals force, and surface tension force. In our dry experiments, $F_{st} \simeq 0$. Variables $d, R_o, R_{dt}, \sigma, \epsilon_0,$ and $A$ represent the distance between the object and the sensor, radii of the object and sensor, charge density, electric constant, and contact area. For flat objects, larger $R_o$ amplifies $F_{vdW}$. The adhesion force allows for grasping when $\Sigma F$ is positive. 

To enhance the adhesion force, we pressed the object against the gel to increase $A$ and minimize $d$ upon contact. The electrostatic aspect can be modified by altering the material of the planar surface, as $\sigma$ is dependent on the object and ground material properties.

To support our assertions, three gel types were created, as shown in Fig. \ref{fig:diffsurface}. The first gel, as shown in Fig. \ref{fig:diffsurface} (a), is made of NOVOCS Gloss and has a transparent surface. To make the second gel, a silicone base mixed with NOVOCS Gloss was initially applied followed by a silicone base with NOVOCS Matte. The applied period ratio between these two materials is $\text{Gloss} : \text{Matte} = 6:4$. The third gel is made of NOVOCS Matte.  The first gel provides the clearest surface, maximizing the contact area during tapping, followed by the second and third gels.

We tested adhesion using a tap grasping method on basil seeds, M1.6 nuts, and M2 nuts. The results were categorized into three categories: no lift, short lift (under 3 seconds), and lift (over 3 seconds) in more than 50 trials each. The outcomes are in Table \ref{table:gelexperiment}.

The clear gel had the highest lifting success rate due to its optimal contact area. While the matte/gloss and matte sensors had notable success with the Basil seed, their performance dropped with the nuts. This reduction is attributed to decreased van der Waals forces since the nuts have a smaller contact area to weight ratio. The basil seed, a non-metallic object, showed dominant electrostatic forces, making lifting easier even without direct contact. This electrostatic effect was amplified when contact was made on surfaces like acrylic as compared to wood and paper. The metallic nuts displayed smaller electrostatic forces because they are highly conductive. Debris can disturb adhesion, therefore periodic cleaning was necessary. These findings aided the decision to use the clear gel in the final sensor design.

\subsection{Grasping thin, small objects via fingernail grasping} \label{sec:strategy}

The fingernail design of the DenseTact-Mini facilitates grasping thin, small objects. Two strategies are proposed for grasping these objects: 1) a rolling contact grasping strategy for objects with sharp edges, and 2) a sliding grasping strategy for objects with rounded edges.
Both strategies employ two DenseTact-Mini units in a two-finger grasping configuration. The first finger uses the DenseTact-Mini with the attached fingernail at its fingertip and dynamically moves when grasping. The second finger utilizes the DenseTact-Mini without the fingernail, exposing just the bare gel, and remains stationary throughout the grasping motion.
\begin{figure}[t]
\vskip 0.04in
\begin{center}
\centerline{\includegraphics[width=0.97\columnwidth]{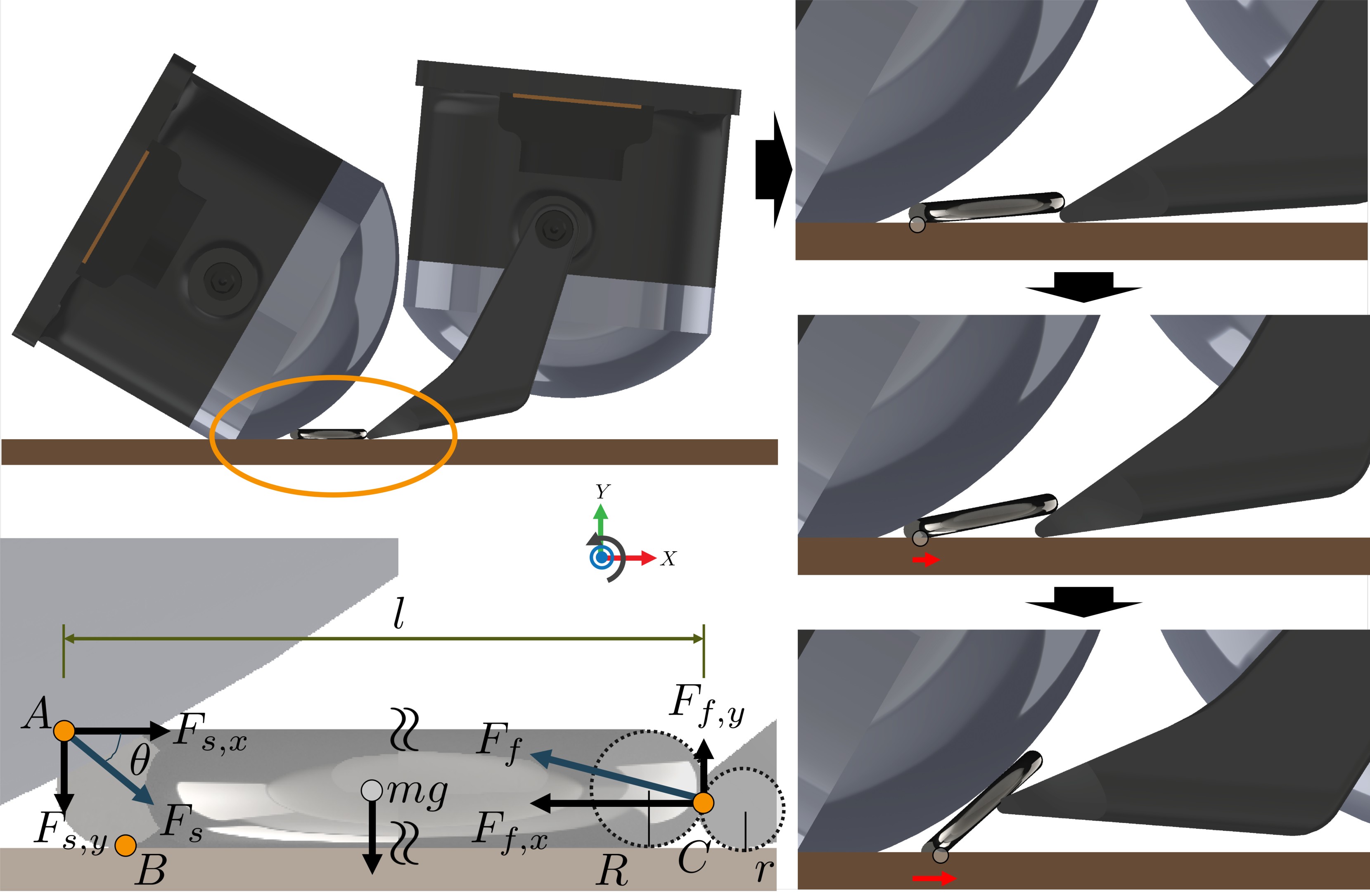}}
\caption{\textbf{Sliding contact strategy.} When the fingernail's radius is smaller than the object, sufficient torque is produced for grasping.}
\label{fig:fn_strat_2}
\end{center}
\vskip -0.4in
\end{figure}
\subsubsection{Rolling contact grasping strategy}
\begin{figure*}[t]
\vskip 0.04in
\begin{center}
\centerline{\includegraphics[width=0.95\textwidth]{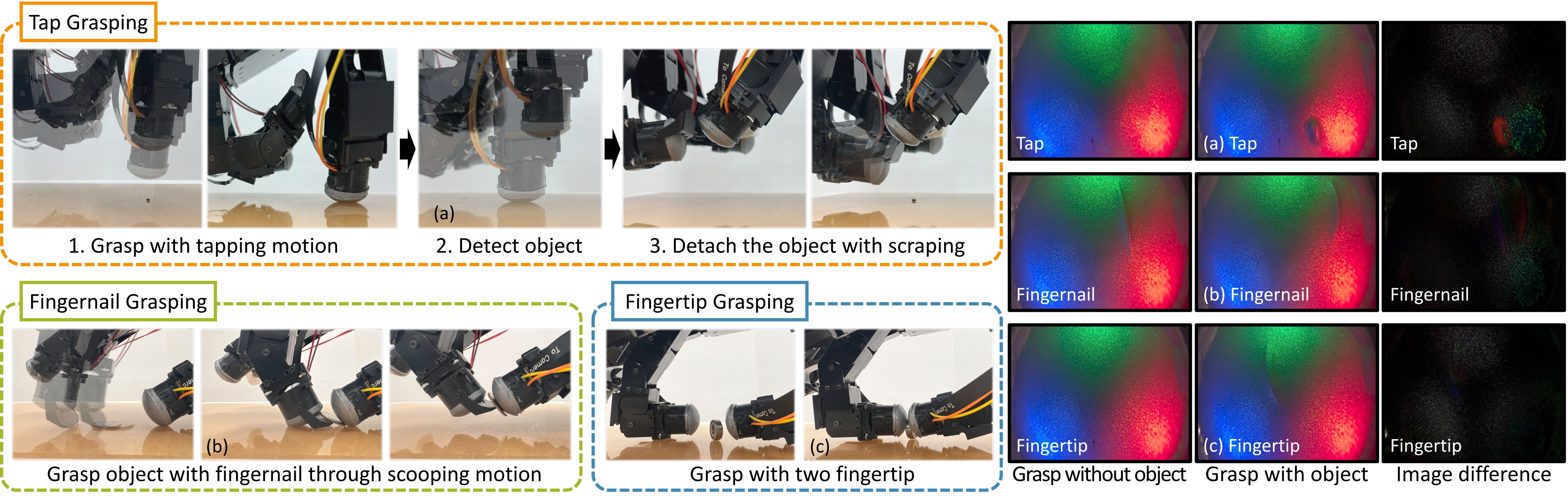}}
\caption{\textbf{Grasping Evaluation Pipeline.} Orange, green, and blue boxes represent the tap, fingernail, and fingertip strategies, respectively. The right-side images depict sensor views with and without objects, and the last column shows image differences for grasp detection.}
\label{fig:pipeline}
\end{center}
\vskip -0.45in
\end{figure*}

Grasping objects with sharp edges using a fingernail results in a different motion compared to grasping objects with rounded edges. The fingernail cannot slide beneath the bottom edge of the object, leading to a rolling contact. Fig. \ref{fig:fn_strat_1} illustrates the schematic of grasping objects with sharp corners or edges. The left bottom image of Fig. \ref{fig:fn_strat_1} presents the free body diagram of the object with a sharp edge. The gel part of the DenseTact-Mini contacts the object at point $A$. The object contacts the ground at point $B$, and the fingernail contacts the object at point $C$. Point $C$ denotes any edge that first contacts between the fingernail and the object, allowing the free-body diagram to be represented in a 2D plane. Although the sensor remains stationary during contact, point $A$ shifts as it makes additional contact due to the gel's hyperelastic properties. Moreover, point $A$ has a higher friction coefficient, $\mu_{g,o}$, since the object is in contact with the silicone, compared to the friction coefficient between the floor surface and the object, $\mu_{s,o}$. Assuming the floor surface is an acrylic board, $\mu_{s,o}$ is similar to the friction coefficient between the fingernail and the object, $\mu_{f,o}$. Given that $\mu_{g,o} >> \mu_{s,o}, \mu_{f,o}$, the object rotates using point $A$ as its pivot. From the free-body diagram, the equations of translational motion, rotational motion, and the force exerted on point $A$ are

\begin{align}
    F_{s,y} &= F_{s,x} \textit{tan} (\theta), \quad \theta = f(p_{obj}, p_{gel}, C_{fem}) \label{eq:theta}\\ 
    \Sigma F_x &= F_{s,x} + R_x - F_{f,x}  =0\label{eq:Fxrolling}\\ 
    \Sigma \tau &=  - \frac{l}{2} mg + h R_x - (h-d)F_{f,x} + lF_{f,y} \label{eq:taurolling}
\end{align}


Where $l,\,h\, ,\,d$ denote the length and height of the object, and the height of the contact between the object and fingernail tip, respectively. The angle $\theta$, analytically defined in Equation \ref{eq:theta}, is a nonlinear function influenced by $p_{obj}, p_{gel}, C_{fem}$, where $p_{obj}, p_{gel}$ indicate the pose and geometrical shape of the object and gel, and $C_{fem}$ references the material property coefficients in hyperelastic material models such as the Yeoh Model \cite{yeoh1993some}.

By substituting equations \ref{eq:theta} and \ref{eq:Fxrolling} into \ref{eq:taurolling}, The equation \ref{eq:taurolling} can be specified in terms of $F_{f,x}$ and $mg$. Then

\begin{align}
   \Sigma \tau &\simeq lF_{f,y}+dF_{f,x} - hF_{s,x} - \frac{l}{2}mg \label{eq:inequality}
\end{align}


 The torque becomes positive if the first two terms of equation \ref{eq:inequality} becomes bigger than the last two terms. Therefore, the object can rotate when influenced by a certain amount of y-directional force while maintaining rolling contact. While not done in this paper, $F_{f,x}, F_{s,x}, $ and $F_{s,y}$ can be measured by estimating force from both DenseTact-Mini sensors through the model proposed in \cite{do2023densetact}.
 
 After the object rotates with point $A$ as the pivot, it slides, causing contact point $B$ to move in the +$x$ direction, as depicted in the right images of Fig. \ref{fig:fn_strat_1}. Once the fingernail successfully positions itself beneath the object's right corner, it establishes sliding contact with the object, resulting in a motion analogous to that described in Section \ref{sec:slidinggrasp}, facilitating the grasp of the object. Throughout this transition, the force $F_{f,x}$ often induces a sudden shift in the object towards the grasp.

\subsubsection{Sliding grasping strategy } \label{sec:slidinggrasp}

When grasping flat, thin objects with rounded edges, or objects with a small gap between their surface and the floor, the fingernail of the DenseTact-Mini aids in the grasp by creating a minor friction point contact between the object and the fingernail. Fig. \ref{fig:fn_strat_1} depicts the overall strategy for grasping rounded-edge objects. The left bottom image displays the free body diagram of the object at the moment of grasping.

If the radius of the right lower edge of the object, $R$, exceeds the edge of the fingernail, $r$, then $F_{f,y}$ generates a positive force, leading to a beneficial torque for grasping. The design file defines $r$ as 0.3mm, with potential deviations due to the 3D printing process. The equation of rotational motion with the dominant term becomes 

\begin{align}
    \Sigma \tau &\simeq  lF_{f,y} - \frac{l}{2}mg >0
\end{align}

Since the position of the fingernail is below the object, $F_{f,y}$ is positive and exceeds the gravitational force. Additionally, such objects typically have a lightweight composition, simplifying the grasp of small, thin items.

\subsection{Grasping larger objects via fingertip grasping}

The DenseTact-Mini, even without the fingernail, can grasp larger objects. With two fingertips as the DenseTact-Mini without fingernails, it is possible to grasp objects of various sizes. Analogous to a two-jaw gripper, the gel portion of the DenseTact-Mini ensures compliant grasping through an expansive area of contact with the object. Following this approach, a fingertip grasp can be characterized as any grasp between two DenseTact-Minis devoid of fingernails. The range of objects graspable, particularly those over 10mm, is influenced by the gripper's specifications, including its DOF and load capacity. The grasping motion and strategy are illustrated in the last row of Fig. \ref{fig:main}. Nonetheless, given the pronounced curvature of the gel, a gripper with a higher DOF or extended link lengths can execute a scooping motion using both fingertips to grasp objects around 10mm, contingent on the object's geometric shape.

\section{Evaluation} \label{sec:evaluation}

\begin{figure}[t]
\vskip 0.04in
\begin{center}
\centerline{\includegraphics[width=0.98\columnwidth]{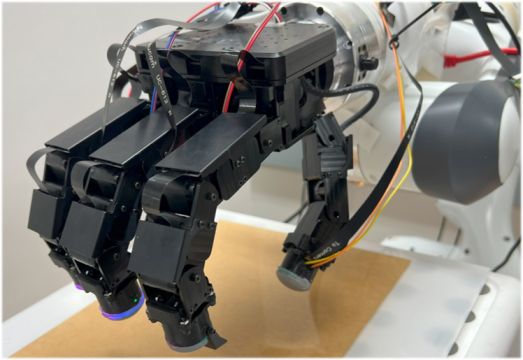}}
\caption{\textbf{Experimental setup}. The Allegro\textsuperscript{TM} hand with DenseTact-Mini, attached to the Franka\textsuperscript{TM} arm, manages object grasping, movement, and detachment.}
\label{fig:experimentsetup}
\end{center}
\vskip -0.4in
\end{figure}
\subsection{Evaluation Pipeline}

In the evaluation pipeline presented in Fig. \ref{fig:pipeline}, different color diagrams (orange, green, blue) represent tap, fingernail, and fingertip grasping strategies. Our experimental setup, detailed in Fig. \ref{fig:experimentsetup}, involves the Allegro\textsuperscript{TM} hand, positioned on the Franka\textsuperscript{TM} robot arm, with DenseTact-Mini sensors on the fingertips. The second and fourth fingers have attached fingernails, while the thumb and third finger, used for tap grasping and fingertip strategies, do not. Although the Jetson Orin can support dual camera sensors, we focused on the DenseTact-Mini thumb sensor for grasp detection. ROS facilitates communication between the components. A state machine oversees the statuses and transitions among `grasp', `move', `ungrasp', and `detect' states, based on the grasping strategy and feedback from the DenseTact-Mini.

In the tap grasping, the process starts with the `grasp' state using a tapping motion. The gripper moves to a set location before transitioning to `detect'. The DenseTact-Mini sensor determines object presence by comparing two baseline images: one with the sensor pressed against an empty surface and another against an object, using pixel intensity variations. Successful grasps advance to the next state, `move', while unsuccessful attempts reset to `grasp'. Once in `ungrasp' after `move', the gripper relocates and tries to release the object by scraping the DenseTact-Mini's gel surface with a fingernail. The fingernail strategy uses a scooping motion, while the fingertip strategy involves the thumb and third finger closing. Both follow the `grasp', `move', and `ungrasp' states. The images on the right side of Fig.\ref{fig:pipeline} depict successful vs. unsuccessful grasp detection. The source code used for this evaluation study is accessible on the project website.

\subsection{Evaluation result}

Table \ref{table:evaluation} presents grasping results for each strategy. Objects are randomly placed within specified regions of interest (ROIs): $10mm \times 10mm$ for tap grasping, $45mm \times 15mm$ for fingernail grasping, and $30mm \times 20mm$ for fingertip grasping. The object's center of mass lies inside the ROI, even if its size exceeds it. Most grasps boast over a 90\% success rate. The M2 nut's heavier weight results in a 90.2\% success rate in tap grasping, while for fingernail grasping, the paperclip, dime, and battery performed best. Failures typically arise when object centers are near the ROI edge or due to unintended rotations. All object images and sizes are available on the project website. Given the large ROIs relative to object sizes, the DenseTact-Mini is suitable for generalized grasping. Allegro\textsuperscript{TM} hand limitations also contribute to grasp failures, suggesting more force could improve grasp outcomes beyond the strategies proposed.

\begin{table}[]
\vskip 0.04in
\begin{tabular}{c|c|c|c|c|}
\cline{2-5}
& \textbf{Object} & \textbf{\begin{tabular}[c]{@{}c@{}}HxLxW\\ (mm)\end{tabular}} & \textbf{\begin{tabular}[c]{@{}c@{}}Weight\\ (g)\end{tabular}} & \textbf{\begin{tabular}[c]{@{}c@{}}Success\\ (Rate (\%))\end{tabular}}  \\ \hline
\multicolumn{1}{|c|}{\multirow{3}{*}{\textbf{\begin{tabular}[c]{@{}c@{}}Tap \\ Grasping\end{tabular}}}}            & Basil Seed      & \begin{tabular}[c]{@{}c@{}}1x1.2x2\end{tabular}
    & 0.0015                                                        & \begin{tabular}[c]{@{}c@{}}52/52 (100)\end{tabular}                                                                 \\ \cline{2-5} 
\multicolumn{1}{|c|}{}                                                                                                    & M1.6 Nut        & \begin{tabular}[c]{@{}c@{}}1.1x3.2x3.2\end{tabular}                                                            & 0.051                                                         & \begin{tabular}[c]{@{}c@{}}48/51 (94.11)\end{tabular}                                                                            \\ \cline{2-5} 
\multicolumn{1}{|c|}{}                                                                                                    & M2 Nut          & \begin{tabular}[c]{@{}c@{}}1.6x3.9x3.9\end{tabular}                                                             & 0.10                                                          & \begin{tabular}[c]{@{}c@{}}46/51 (90.2)\end{tabular}                                                                              \\ \hline
\multicolumn{1}{|c|}{\multirow{5}{*}{\textbf{\begin{tabular}[c]{@{}c@{}}Fingernail \\ grasping\end{tabular}}}} & Paperclip       & 
\begin{tabular}[c]{@{}c@{}}0.8x6.9x26.5\end{tabular}                                                          & 0.31                                                          & \begin{tabular}[c]{@{}c@{}}48/51 (94.12)\end{tabular}                                                                             \\ \cline{2-5} 
\multicolumn{1}{|c|}{}                                                                                                    & \begin{tabular}[c]{@{}c@{}}Small\\Wrench\end{tabular}  & \begin{tabular}[c]{@{}c@{}}0.9x7x45\end{tabular}                                                          & 1.95                                                          & \begin{tabular}[c]{@{}c@{}}46/51 (90.2)\end{tabular}                                                                            \\ \cline{2-5} 
\multicolumn{1}{|c|}{}                                                                                                    & Dime            & \begin{tabular}[c]{@{}c@{}}1.3x17.9x17.9\end{tabular}                                                         & 2.24                                                          & \begin{tabular}[c]{@{}c@{}}54/57 (94.74)\end{tabular}                                                                           \\ \cline{2-5} 
\multicolumn{1}{|c|}{}                                                                                                    & \begin{tabular}[c]{@{}c@{}}CR2032\\Battery\end{tabular}  & \begin{tabular}[c]{@{}c@{}}3.2x20x20\end{tabular}                                                        & 3.00                                                          & \begin{tabular}[c]{@{}c@{}}50/53 (94.34)\end{tabular}                                                                          \\ \hline
\multicolumn{1}{|c|}{\textbf{\begin{tabular}[c]{@{}c@{}}Fingertip \\ grasping\end{tabular}}}                  & Bearing         & \begin{tabular}[c]{@{}c@{}}16x5x16\end{tabular}                                                          & 4.57                                                          & \begin{tabular}[c]{@{}c@{}}50/53 (94.34)\end{tabular}                                                                        \\ \hline
\end{tabular}
\caption{\textbf{Grasping success rate for each strategy.} The success rate for all daily objects of various sizes and weights is greater than 90\%.}
\label{table:evaluation}
\vskip -0.3in

\end{table}

\section{Conclusions}\label{sec:conclusion}

This paper presents three distinct grasping strategies for handling various small objects from flat surfaces using the miniaturized vision-based tactile sensor, the DenseTact-Mini. The DenseTact-Mini comprises of a high-resolution soft gel component with a modular stacked design, a smooth gel surface, and a detachable fingernail design. These attributes facilitate three varied grasping techniques: 1) adhesive force-based tapping for objects sized between 1mm and 3mm, 2) scooping motion with fingernail for grasping thin and small objects, and 3) a conventional two-fingertip grasp for objects larger than 10mm.
Through evaluating everyday small objects, including tiny nuts and slender items, we have demonstrated that our sensor, coupled with the fingernail and an appropriate multi-DOF gripper, can effectively grasp a variety of multi-sized objects by applying different grasping strategies. The success rate exceeded 90\% for all evaluated objects. In future work, we aim to extend the length of each finger's linkage on the gripper, assess dexterous manipulation capabilities using the DenseTact-Mini sensor, and calibrate the sensor for precise grasp manipulation. We believe the DenseTact-Mini is pivotal for planning multi-object grasps and intricate in-hand manipulation with tactile sensing image input.

\addtolength{\textheight}{-5cm}   









  \bibliographystyle{./IEEEtran} 
  \bibliography{./IEEEexample}

\end{document}